\documentclass{article}
\usepackage{ijcai26}

\usepackage{times}
\usepackage{soul}
\usepackage{url}
\usepackage[hidelinks]{hyperref}
\usepackage[utf8]{inputenc}
\usepackage[small]{caption}
\usepackage{graphicx}
\usepackage{amsmath}
\usepackage{amsthm}
\usepackage{booktabs}
\usepackage{algorithm}
\usepackage{algorithmic}
\usepackage[nolist]{acronym}
\usepackage{xcolor}
\usepackage{subcaption}
\usepackage{listings}
\usepackage[T1]{fontenc}
\usepackage{inconsolata} % bold monospace works well
\usepackage{comment}
\usepackage{makecell}
\usepackage{algorithm}
\title{Implementing Knowledge Representation and Reasoning with Object Oriented Design}

\author{
Abdelrhman Bassiouny\and
Tom Schierenbeck\and
Sorin Arion\and
Benjamin Alt\and
Naren Vasantakumaar\and
Giang Nguyen\And
Michael Beetz\\
\affiliations
AICOR Institute for Artificial Intelligence\\
University of Bremen\\
Bremen, Germany\\
\emails
bassioun@uni-bremen.de,
tom\_sch@uni-bremen.de,
sorin@uni-bremen.de,
benjamin.alt@uni-bremen.de,
naren@uni-bremen.de,
hoanggia@uni-bremen.de,
beetz@cs.uni-bremen.de
}

\begin{document}

% Define acronyms
\newacro{krr}[KR\&R]{knowledge representation and reasoning}
\newacro{eql}[EQL]{Entity Query Language}
\newacro{oom}[OOM]{object-ontological mapping}
\newacro{orm}[ORM]{object-relational mapping}
\newacro{oop}[OOP]{object-oriented programming}
\newacro{ood}[OOD]{object-oriented design}
\newacro{vr}[VR]{virtual reality}
\newacro{rdr}[RDR]{Ripple Down Rule}
\newacro{scrdr}[SCRDR]{Single Class Ripple Down Rules}
\newacro{mcrdr}[SCRDR]{Multi-Class Ripple Down Rules}
\newacro{grdr}[GRDR]{General Ripple Down Rules}
\newacro{cbr}[CBR]{case-based reasoning}
\newacro{krrood}[KRROOD]{Knowledge Representation and Reasoning with Object Oriented Design}
\newacro{dl}[DL]{description logic}
\newacro{ai}[AI]{artificial intelligence}
\newacro{dao}[DAO]{Data Access Object}
\newacro{fol}[FOL]{First Order Logic}
\newacro{owl}[OWL]{Web Ontology Language}
\newacro{rdf}[RDF]{Resource Description Framework}
\newacro{ml}[ML]{machine learning}

\lstdefinestyle{clean}{
  language=Python,
  basicstyle=\ttfamily\small,
  keywordstyle=\color{blue}\bfseries,     % Python keywords
  stringstyle=\color{orange},
  commentstyle=\color{gray}\itshape,
  numbers=left,
  numberstyle=\tiny\color{gray},
  frame=single,
  breaklines=true,
  tabsize=2
}

\maketitle

\begin{abstract}
This paper introduces KRROOD, a framework designed to bridge the integration gap between modern software engineering and \ac{krr} systems. While \ac{oop} is the standard for developing complex applications, existing \ac{krr} frameworks often rely on external ontologies and specialized languages that are difficult to integrate with imperative code. KRROOD addresses this by treating knowledge as a first-class programming abstraction using native class structures, bridging the gap between the logic programming and OOP paradigms. We evaluate the system on the OWL2Bench benchmark and a human-robot task learning scenario. Experimental results show that KRROOD achieves strong performance while supporting the expressive reasoning required for real-world autonomous systems.
\end{abstract}

\section{Introduction}
\label{sec:introduction}

Intelligent autonomous systems benefit from internal models to interpret their environment, reason about task requirements, and select actions that robustly achieve goals. The \acf{krr} paradigm provides interpretable structures and inference mechanisms that support generalization across tasks and domains. Decades of research demonstrate that explicit knowledge models enable tractable problem solving through abstraction, modularity and explanation at the knowledge level \cite{delgrande2024current,newell1982knowledge}. Yet for roboticists and domain experts deploying intelligent systems in real-world settings, the practical adoption of \ac{krr} depends not only on representational adequacy but on the seamless integration of knowledge structures with perception, planning, control and application logic \cite{toberg2024commonsense}.

This integration challenge exposes a gap between two programming paradigms. The dominant paradigm for building complex software systems is \acf{oop}, which provides modularity, data validation, and scalable development practices. In contrast, \ac{krr} systems typically rely on logic programming, and operate under fundamentally different assumptions about completeness, openness and execution semantics. As a consequence, developers must maintain two separate representational systems that do not share execution models or tooling ecosystems \cite{ledvinka2020comparison}, leading to an object-ontological equivalent of the Object-Relational Impedance Mismatch \cite{ireland2015exposing}. This mismatch is particularly acute in Python-based \ac{ai} systems, where perception, planning, and learning APIs increasingly converge, but no actively maintained \ac{krr} framework provides native, object-oriented integration of both knowledge representation and advanced reasoning capabilities \cite{abicht2023owl}.

\begin{figure}[t]
  \centering
  \includegraphics[width=\columnwidth]{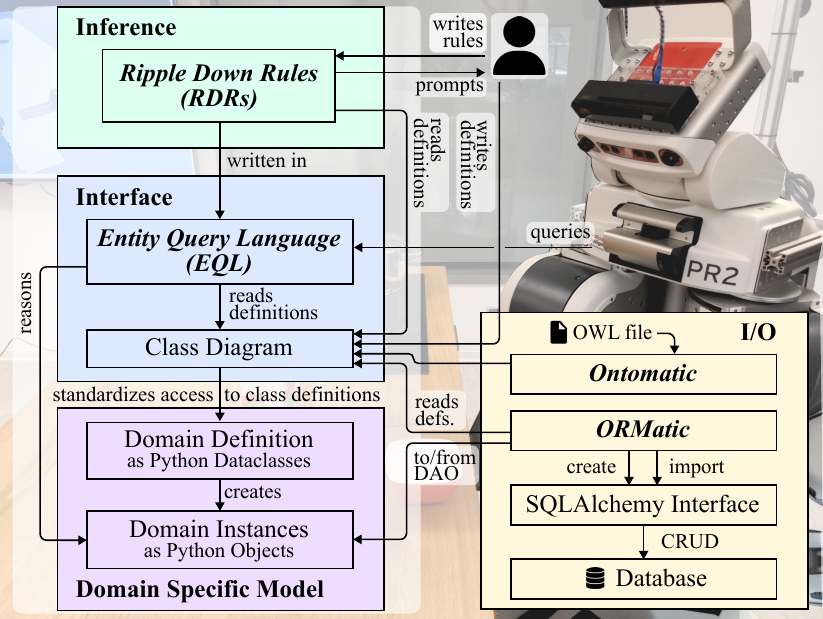}
  \caption{The \acs{krrood} framework provides representations and tooling for native, object-oriented \acs{krr} in Python.}
  \label{fig:arch}
\end{figure}

Beyond being an engineering inconvenience, this mismatch reflects a deeper tension in knowledge systems research. Frames \cite{minsky1974framework}, scripts \cite{schank1977scripts}, and languages such as FRL \cite{roberts1977frl} and KRL \cite{bobrow1977overview} were designed to support reasoning over structured objects. Modern \ac{oop} inherits this tradition but has diverged from modern \ac{krr} tooling, which has come to increasingly rely on external reasoners and \ac{dl} formalisms that treat knowledge as a resource separate from application logic \cite{buoncompagni2024owloop}.

This paper argues that bridging this paradigm gap requires treating knowledge as a \emph{first-class programming abstraction} rather than an external resource. It introduces \acf{krrood}, a framework that unifies \ac{krr} capabilities within an \ac{oop} Python architecture and makes expressive \ac{krr} accessible within standard software development workflows. We make five core contributions:

\begin{enumerate}
\item An interpretation of Python datastructures as knowledge representation that treats domain entities as structured objects directly accessible to application code.

\item \acf{eql}, a declarative querying language for Python data structures that extends upon well-established conjunctive query languages while remaining tractable and decidable.

\item \Acp{rdr} for consistent incremental knowledge update and maintenance through expert interaction during system runtime.

\item ORMatic, a persistence layer that transparently stores, indexes, and retrieves knowledge objects while maintaining alignment between domain models and data schemas.

\item Ontomatic, a tool for automatic migration of knowledge represented in the \ac{owl} into object-oriented Python data structures.
\end{enumerate}

Taken together, they bridge the paradigm gap between application programming and \ac{krr} by enabling engineers to seamlessly \textit{program with knowledge}.
We validate the capabilities and performance of \ac{krrood} on OWL2Bench \cite{singh2020owl2bench} and a human-robot task learning scenario. \ac{krrood} is fully open-source\footnote{Source code: \url{https://github.com/cram2/cognitive_robot_abstract_machine}} and integrated into a robot cognitive architecture \cite{beetz2025robot}, providing foundational capabilities for hybrid physical \ac{ai}.

\section{Related Work}
\label{sec:related-work}

The mismatch between the representations used in \ac{krr} and those used in application programming has been identified as a core challenge in knowledge engineering \cite{ledvinka2020comparison,baset2018object,banse2024owl2proto}. \Ac{oom} systems address the structural correspondence between application objects and their representations in knowledge bases. In Java, JOPA \cite{ledvinka2016jopa}, Jastor \cite{szekely2009jastor}, KOMMA \cite{wenzel2010komma}, and the OWLAPI \cite{horridge2011owl} offer programmatic ontology access with varying degrees of object integration. Owlready2 \cite{lamy2017owlready} is a Python equivalent. More recent frameworks such as OWLOOP \cite{buoncompagni2024owloop} extend this line of work by mapping OWL axioms into \ac{oop} hierarchies with polymorphism support, allow downstream applications to use ontologies in protocol buffers \cite{banse2024owl2proto} or project ontology information into vector spaces for downstream \ac{ml} models \cite{zhapa2023mowl}. These frameworks facilitate programmatic ontology access, but delegate inference to external reasoners and do not support rule-based reasoning.

External reasoners such as RacerPro \cite{haarslev2012racerpro}, HermiT \cite{glimm2014hermit}, RDFox \cite{nenov2015rdfox} and MORe \cite{armas2012more} provide sound and complete inference for various \ac{dl} fragments. Most modern \ac{krr} systems such as DeepOnto \cite{he2024deeponto} rely on reasoning as a batch process over serialized ontology files: applications must export domain state, invoke the reasoner and parse results. This separation forces developers to maintain parallel representations of domain knowledge and synchronize state across system boundaries, complicating development and maintenance. KERAIA \cite{varey2025keraia} is a recent self-contained knowledge system with dynamic aggregation and context-sensitive inheritance. However, it introduces its own representation language (KSYNTH), and integrating it into \ac{oop} applications reintroduces the impedance mismatch at the system boundary.

\subsection{KRR as a Native Programming Abstraction}

The vision of treating knowledge as a native programming construct has historical roots in frames \cite{minsky1974framework}, FRL \cite{roberts1977frl}, and KRL \cite{bobrow1977overview}, which were designed to support reasoning over structured objects. Systems such as ErgoAI \cite{kifer2018ergoai} and KnowRob \cite{beetz2018know} realized this vision by providing built-in reasoning through Prolog's unification mechanism. However, they also reintroduce the impedance mismatch at the system boundary, requiring developers to context-switch between paradigms. Logic programming libraries for Python embed Datalog or probabilistic inference, but do not provide object-oriented knowledge representation \cite{carbonnelle2020pydatalog,dries2015problog2}.

OWLAPY \cite{baci2025owlapy} is perhaps closest in spirit to \ac{krrood}: it models OWL entities as first-class Python objects and provides both Python-native and external reasoners. However, reasoning operates over \ac{owl} constructs rather than application domain objects, requiring developers to maintain a mapping between the two. OWLAPY also does not support rule-based reasoning or integrated persistence. 

No existing framework unifies object-oriented representation, expressive querying, native rule-based reasoning and persistence within a single abstraction. \Ac{krrood} addresses this gap by treating knowledge as a first-class programming construct accessible via native Python mechanisms.

\section{EQL: A Python-Based Query and Logical Axiomatization Language}
\label{sec:eql}

We address the paradigm gap by proposing an object-oriented, Python-native representation of knowledge. In \ac{krrood}, concepts are represented as classes, relations between concepts are captured by class attributes and n-ary functions, and taxonomic relations are modeled through inheritance. Although this representation differs structurally from First-Order Logic (FOL), for example, a Python class definition does not explicitly encode axioms, it is sufficiently flexible to allow arbitrary class methods that express statements holding generally for the class, as demonstrated by the experiments in Section \ref{sec:experiments}.
Queries over this knowledge can be formulated over classes, their instances, and their attributes.

To make this object-oriented knowledge representation practically usable as a \ac{krr} system, a mechanism is required that allows structured access to classes, instances, and their relations, while preserving the flexibility of the underlying Python execution model. This motivates a native query language that supports both declarative retrieval and procedural reasoning. We propose \ac{eql}, a declarative query interface for knowledge retrieval and axiomatization within Python applications. %Unlike external query languages (e.g., SPARQL, Prolog), \ac{eql} operates directly over domain objects and supports the use of arbitrary Python procedures within query predicates. This enables domain-specific reasoning, such as kinematic computations in robotics, to be integrated seamlessly into logical inference without external reasoners or language bindings.

\begin{comment}
It follows three main design principles:
% TOM i dont like the bullet point like writing
\textbf{Python-native querying}: \ac{eql} offers a declarative syntax layered on top of Python, allowing queries to be expressed and executed directly on Python data structures.

\textbf{Domain-aware reasoning}: Queries make use of the data structures and procedures of the application domain. In robotics settings, for example, geometric reasoning, kinematics, or physics simulations can be evaluated as part of a query.

\textbf{Built-in reasoning}: \ac{eql} includes a built-in constrained version of backward chaining to ensure decideability and real-time performance. This removes the need for external reasoners while enabling inspection of queries and reasoning routines at runtime.
% TOM i would say inspection is only one aspect of what it enables

% Rather than aiming to replace existing logic formalisms, EQL is designed to support practical reasoning over explicitly represented entities and relations in finite, dynamically evolving environments.

\end{comment}

\subsection{Grammar and Example Queries}
\label{ssec:example_queries}

The syntax of \ac{eql} follows in spirit conjunctive database query languages with a Select-Where-From pattern, but adds declarative syntactical elements to facilitate \ac{krr} applications. \ac{eql} statements are \textit{descriptions of entities}, rather than imperative programming commands (see Listing \ref{lst:simple}).

Similar to tables in SQL and nodes/edges in graph query languages, \ac{eql} has \textit{variables} as the atomic data holders. Variables are symbolic representations of Python objects that represent a set of values. They can be an arbitrary set of values or values of a specific class. Attributes of objects represented by variables can be accessed exactly as they are accessed on the objects themselves, but return symbolic representations of the attributes rather than their concrete value.

An \ac{eql} variable is constructed from a \textit{type} (a class), a \textit{domain} (a set of values), or both:
\begin{center}
\texttt{var = variable(Type, Domain)}
\end{center}
Explicit domain specification enables queries to operate over program-computed sets, e. g., sets produced by methods of an object, rather than exhaustive class extensions, distinguishing \ac{eql} from traditional database query languages.
This design permits the integration of domain-specific computation into query evaluation. Variables in \ac{eql} represent symbolic constraints over value sets, while program variables may be used directly as literals when denoting constants.

An \ac{eql} query follows the structure:
\begin{center}
\texttt{query = ResultProcessor(EntityDescriptor)}
\end{center}
The \textit{result processor} applies either \textit{aggregation} (e.g., \texttt{sum}, \texttt{count}) or \textit{quantification} to the retrieved entities. Quantifiers constrain the cardinality of results: \texttt{a}/\texttt{an} permits zero or more matching entities (existential quantification), while \texttt{the} enforces uniqueness.

An \textit{entity descriptor} specifies retrieval constraints analogous to the Select-Where clause in SQL. The descriptor \texttt{entity(variable)} retrieves a single entity, while \texttt{set\_of(var1, var2, ...)} retrieves entity tuples. Constraints are specified through \texttt{where(cond1, cond2, ...)}, which accepts conjunctive conditions over entity attributes.

\begin{comment}
The following example queries illustrate the expressive power of \ac{eql}. Given the corresponding domain object implementations, they are executable as Python code.
\end{comment}

Listing~\ref{lst:simple} finds all persons of age 20:
\begin{lstlisting}[
  style=clean,
  caption={A simple entity query},
  label={lst:simple},
  emph={an,where},
  emphstyle=\color{blue}\bfseries,
  emph={[2]entity,contains,variable},
  emphstyle={[2]\color{purple}},
  emph={[3]Person},
  emphstyle={[3]\color{teal}\bfseries},
  emph={[4]domain},
  emphstyle={[4]\color{brown}}
]
p = variable(Person)
query = an(entity(p).where(p.age == 20))
\end{lstlisting}

%TODO: Mention that an() yields a list/iterator/? of Person objects.
%TOOD: Why is there always domain=None? If domain is default-initialized to none anyway, omit it. It looks confusing and would require evaluation.

% Listing~\ref{lst:match} shows a corresponding higher-level query using the \texttt{match\_variable} \ac{eql} construct for structure pattern matching, such that in the same statement where a variable is defined, its attributes can be constrained:

% \begin{ting}[
%   style=clean,
%   language=Python,
%   caption={An entity query with pattern matching},
%   label={lst:match},
%   emph={an,where},
%   emphstyle=\color{blue}\bfseries,
%   emph={[2]entity,contains,variable,match_variable},
%   emphstyle={[2]\color{purple}},
%   emph={[3]Person},
%   emphstyle={[3]\color{teal}\bfseries},
%   emph={[4]age,domain},
%   emphstyle={[4]\color{brown}}
% ]
% an(entity(match_variable(Person, domain=None)(age=20)))
% \end{ting}

Listing~\ref{lst:complicated-query} shows a nested query with domain specification, quantification (universal and existential), multiple selected variables, and indexing of collections. The query selects robot and capability pairs where the robot possesses the capability, satisfies part-size constraints, and has sufficiently dexterous hands that have at least 5 fingers; variable definitions are omitted.

\begin{lstlisting}[
  style=clean,
  language=Python,
  caption={A more complex query},
  label={lst:complicated-query},
  % DSL keywords / quantifiers
  emph={a,where,exists,for_all},
  emphstyle=\color{blue}\bfseries,
  % DSL functions/predicates
  emph={[2]set_of,contains,count},
  emphstyle={[2]\color{purple}},
]
query = a(set_of(robot, capability).where(
   contains(robot.capabilities, capability),
   robot.parts.size[0] <= 1,
   for_all(arm, count(arm.fingers) >= 5))))
\end{lstlisting}

\subsection{Semantics of \ac{eql}}
EQL is grounded in the well-established framework of conjunctive queries, which correspond to the existential, conjunction-only fragment of \ac{fol}.

To increase expressiveness while retaining executability, \ac{eql} extends conjunctive queries with three carefully chosen features. First, \ac{eql} supports \textbf{unions of conjunctive queries}, allowing disjunctive pattern matching (e.g., querying entities that satisfy one of several alternative conditions). Second, it adopts \textbf{negation as failure}, interpreting negation operationally as the absence of derivable evidence in the knowledge base. Third, it provides a \textbf{restricted form of universal quantification}, interpreted over the active domain of entities currently present in the knowledge base.

These choices reflect the closed-world assumptions commonly adopted in databases and robotic world models, where the knowledge base represents the agent’s current belief state. While they are not as expressive as full \ac{fol}, they preserve decidability, tractability, and align the semantics of the language with the requirements of real-world \ac{ai} systems.

\section{Python Native Rule Based Reasoning}
\label{sec:rdrs}
Rule-based reasoning has been used extensively in domains that require correctness, explainability, and extensibility. The advantages of rule-based systems come at the cost of ease of maintenance: As the size of the knowledge base increases, the probability of conflict between rules also increases \cite{compton2021rdr}. A range of conflict-resolution strategies have been proposed, including specificity- and recency-based rule selection \cite{brachman2004knowledge}. However, these strategies are inherently heuristic: while they may reduce the likelihood of rule conflicts, they do not eliminate them.

\ac{rdr}s \cite{compton2021rdr} offer a pragmatic solution by building the knowledge base rule by rule from available data using expert supervision. While rules are being added, conflicts are automatically detected and the expert is prompted to resolve them by answering one basic question: What are the conditions that differentiate these two conflicting rules? The system then automatically places the answer as a refinement rule in the appropriate place in the rule tree.

There are three established types of \ac{rdr}s: SingleClassRDRs infer concetps that are mutually exclusive; MultiClassRDRs infer non mutually exclusive concepts; and GeneralRDRs, which can contain multiple different \ac{rdr} trees for the inference of different concepts. \ac{rdr} trees are run until no extra inference is produced, making use of intermediate inferences for further activation of rules that may use them, thus inferring higher level concepts from them.

\ac{krrood} contributes a Python-native \ac{rdr} implementation that makes use of object-oriented design and the Python ecosystem. Additionally, it extends \acp{rdr} with features that increase its maintainability and expressiveness.

\subsection{RDR Extensions in KRROOD}
The \ac{krrood} \ac{rdr} implementation provides three key extensions to the traditional formalism:

\textbf{Flexible case representation.} Cases may be arbitrary Python objects, enabling direct integration with \ac{oop} domain models. Python functions can be annotated to define \acp{rdr}, with function inputs as case descriptions and outputs as inferred conclusions.

\textbf{Expressive rule conditions.} Rule conditions are executable Python code rather than a specialized language. This enables direct use of domain-specific APIs and eliminates knowledge duplication across formalisms.

\textbf{Maintainable rule representation.} Rule bases are materialized as auto-generated Python modules comprising both the tree structure and logic. These modules are independently executable and support direct user modification, enabling maintenance as domain models evolve within standard development workflows.

\textbf{Interactive rule writing.} The \ac{rdr} prompt opens an editor with template code of a function that takes the case as input and expects a conclusion of the given type as output. The expert then writes the rule in the function body, utilizing the full capabilities of the Python language and leveraging IDE support.

In addition, \ac{rdr} trees can be directly defined using \ac{eql}. In Listing~\ref{lst:1_rule_tree}, the initial tree contains a single rule that infers that the parent body of a fixed connection is a \textit{Drawer} when the child body of the connection is of type \textit{Handle}. When a new case matches this rule but corresponds to a \textit{Door} rather than a \textit{Drawer}, \ac{rdr} requests a distinguishing condition from the expert. The expert specifies that a \textit{Door} has a body larger than $1\,m$. The rule tree is then automatically refined by adding this condition as a child rule, which overrides the original conclusion and infers a \textit{Door} when the refinement condition holds.

\begin{lstlisting}[
  style=clean,
  caption={RDR tree with Refinement using EQL},
  label={lst:1_rule_tree},
  emph={Add},
  emphstyle=\color{blue}\bfseries,
  emph={[2]HasType, inference, refinement, variable},
  emphstyle={[2]\color{purple}},
  emph={[3]Drawer,Door},
  emphstyle={[3]\color{teal}\bfseries},
  emph={[4]with},
  emphstyle={[4]\color{brown}}
]
fixed_connection = variable(FixedConnection)
child = fixed_connection.child
parent = fixed_connection.parent
with HasType(child, Handle):
  Add(inference(Drawer)(container=parent))
  with refinement(parent.size > 1):
    Add(inference(Door)(body=parent))
\end{lstlisting}

\section{Ontomatic: OWL Conversion}
\label{sec:ontomatic}

\ac{krrood} provides tool support for the automatic conversion of OWL representations into \ac{ood} data structures. Ontomatic, the OWL adapter of \ac{krrood}, converts \ac{owl} ontologies in two phases. The first phase involves generating the classes and properties to represent the T-Box, including inference of subsumption between classes, inference of domains and ranges of properties, and conversion of \ac{owl} axioms into an equivalent \ac{eql} representation. Two prominent challenges are addressed in this phase:

\textbf{Non-disjoint sibling classes.} In \ac{owl}, sibling classes are not necessarily disjoint, allowing an individual to belong to multiple sibling classes simultaneously. This contrasts with object-oriented inheritance. 

To bridge this mismatch, we model overlapping class membership using a Role pattern: a persistent entity represents the individual’s identity (URI), and role-specific classes reference it via composition rather than inheritance. This allows an individual to assume multiple roles concurrently, in line with OntoClean modeling principles \cite{Guarino2004}. Roles can either be explicitly defined in the ontology by adding a \texttt{roleFor} property, or they can be implied through disjointness constraints.

\textbf{Expressing axioms.} \ac{owl} axioms expressed as class descriptions and restrictions cannot be fully captured through class structures alone, as inheritance hierarchies encode only taxonomic relationships and attribute ranges rather than complex constraints involving chained attributes or quantified conditions. We address this limitation by encoding axioms as class-level methods that implement verification predicates using \ac{eql}. These predicates operate over arbitrary domain instances and support existential and universal quantification. Listing~\ref{lst:axiom} illustrates the \ac{ood} representation of the constraint from OWL2Bench (DL profile) that a \texttt{LeisureStudent} is a student that takes at most one course. This is represented in OWL as an intersection of subclass of \textit{Student}, and an \textit{onProperty} restriction on the \textit{takesCourse} property with a \textit{maxQualifiedCardinality} of 1 of class \textit{Course}.

\begin{lstlisting}[
  style=clean,
  caption={Class with EQL Axiom},
  label={lst:axiom},
  emph={exists,count},
  emphstyle=\color{blue}\bfseries,
  emph={[2]IsSubClassOrRole,HasProperty},
  emphstyle={[2]\color{purple}},
  emph={[3]Student, Course, TakesCourse, LeisureStudent},
  emphstyle={[3]\color{teal}\bfseries},
  emph={[4]@classmethod},
  emphstyle={[4]\color{brown}}
]
class LeisureStudent(Student):
 @classmethod
 def axiom(cls, candidate):
  return (
   exists(IsSubClassOrRole(candidate.types, Student)),
   HasProperty(candidate, TakesCourse),
   count(IsSubClassOrRole(candidate.takes_course.types, Course)) <= 1
   )
\end{lstlisting}

In a second phase, Ontomatic loads individuals as instances of the generated classes. The appropriate types for each instance are inferred using explicitly declared types, assigned properties and class axioms.

To improve performance, on-time forward chaining is applied to all properties except those that are both transitive and symmetric, as handling these properties eagerly leads to a substantial increase in reasoning time (from ~15 seconds to around 100 minutes in our experiments). To preserve completeness, a final post-processing step extracts the induced subgraphs of symmetric-transitive properties and computes their weakly connected components. Within each component, all node pairs are inferred to be related, thereby recovering the missing relations and completing the inference. 

\section{ORMatic: Object-Relational Mapping and Persistence}
\label{sec:ormatic}

\Ac{krr} applications require robust persistence solutions to manage the complex, evolving states of domain knowledge. While Python dataclasses offer a high-fidelity representation of structured data, the challenge extends beyond simple integration and involves maintaining referential integrity and schema synchronization across the application lifecycle. Without a dedicated abstraction layer, systems risk ``schema drift'' and the fragmentation of domain logic. \Ac{orm} addresses these risks by providing a declarative framework that enforces data constraints, manages complex relational networks, and ensures the persistent state's alignment with the application knowledge model.

The design of ORMatic, \ac{krrood}'s persistance layer, is guided by several core principles intended to streamline the development of knowledge-based systems. Primarily, the system emphasizes maintainability and scalability, ensuring the persistence layers ability to grow alongside the application. ORMatic was designed with \textbf{
\ac{ml} compatibility} in mind: A modern \ac{krr} system should allow for direct use of its data for machine learning, which most often requires tabular data as input \cite{kleppmann2019designing}.
Another critical requirement is \textbf{non-interference}: the tool should not dictate the structure of domain-specific logic \cite{davis1993knowledge}. By minimizing the object-relational impedance mismatch \cite{neward2006vietnam}, ORMatic allows developers and \ac{ai} agents to use standard Python \ac{oop} patterns.
A further requirement is the strict alignment with SOLID principles, particularly the \textbf{Single Responsibility Principle (SRP)}. In many traditional \ac{orm} implementations, the failure to separate the \acp{dao} from the domain class forces domain objects to handle their own persistence routines. This violation of SRP often introduces side effects, that interfere with domain-specific logic, slowing down execution and causing unexpected behavior. ORMatic avoids this by maintaining a clean separation between the logic used for computation and the structures used for storage.

ORMatic provides sensible defaults while remaining extensible through customizable mappings and supports a variety of database backends to suit different infrastructure needs. These requirements were discovered in episodic memory components for cognitive architectures \cite{beetz2025robot}.
In complex software systems, specific parts of the domain model will likely require persistence logic or conversion routines that differ from the automatically generated defaults. 
ORMatic provides a structured framework to define alternative persistence specifications or specialized serialization routines. These custom definitions are integrated into the global persistence pipeline, ensuring that developers can achieve full customization while still using dataclasses as the primary framework for defining information.

ORMatic functions as a translation and generation layer between Python domain models and relational storage. The system automatically generates an SQLAlchemy interface that mirrors the structure and behavior of existing Python classes. To facilitate data movement, it provides built-in routines to convert domain objects into \acp{dao} and back automatically.
This is different to many existing solutions that require extra definitions for data schemas and validation, (e. g. SQLAlchemy\footnote{\url{https://www.sqlalchemy.org/}}, Pydantic\footnote{\url{https://docs.pydantic.dev/latest/}}, peewee\footnote{\url{https://docs.peewee-orm.com/en/latest/index.html}}). ORMatic operates entirely within the application's existing dataclasses (see Fig. \ref{fig:arch}). 
This approach ensures that the data model and the domain model remain identical, reducing the cognitive load on the developer and eliminating synchronization errors. By utilizing SQL as a backend, ORMatic benefits from the reliability and query capabilities of established relational database engines.

%\subsection{Core Contribution}

\section{Experiments}
\label{sec:experiments}

%- We want to show that 
 %   - we can do the same stuff as Ontology based KRR by using OOP. 
  %  - We are able to give the same answers to the same questions asked by KRR systems
   % - We are able to reason in the same conclusions as the OWL stuff does, but in OOP
    %- We are competitive regarding the reasoning and querying time, even without owl
    %- The KRROOD approach scales to robotics framework where KRR can be used to introspect robotic behavior
This chapter evaluates \ac{krrood} as an integrated \ac{krr} system on OWL2Bench and a robotic task learning application. The experiments show that \ac{dl} statements can be expressed directly on object-oriented knowledge structures, preserving ontological semantics while operating entirely within an \ac{ood} paradigm. The evaluation, therefore, addresses not only expressivity but also whether such an approach is competitive in terms of loading, reasoning, and querying performance.
    
\subsection{OWL2Bench Benchmarks}

We evaluate \ac{krrood} on the OWL2Bench benchmark with the OWL 2 RL profile \cite{singh2020owl2bench}.\footnote{ The experiments can be reproduced at \url{https://github.com/tomsch420/krrood_experiments}.}
The experimental setup utilizes a machine running Ubuntu 24.04.3 LTS, equipped with a 11th Gen Intel® Core™ i7-11700k processor and 32 GB of RAM.

An object-oriented model (referenced as \ac{krrood} in Tables \ref{tab:baselines} and \ref{tab:queries}) is auto-generated from the OWL2Bench ontology using Ontomatic, with only minor changes related to the Role pattern: A few classes were modified to explicitly mention that they are roles and not just subclasses, to handle the issues arising from (non) disjoint sibling classes discussed in Section \ref{sec:ontomatic}. From the generated classes, an \ac{orm} interface is automatically derived using ORMatic and the data is persisted in a PostgreSQL database. To assess the impact of materializing object-oriented data structures on query performance, we include SQLAlchemy on that PostgreSQL data as a baseline in the query experiment (Table \ref{tab:queries}). SQLAlchemy is not compared in reasoning (Table \ref{tab:baselines}), as it does not support reasoning on its own. 

\subsubsection{Loading and Reasoning}

Loading and reasoning performance is compared  across six systems: KRROOD, owlready2 with Pellet, Protégé with Pellet, GraphDB with OWL RL (Optimized), and RDFlib with owlrl.

Performance is measured for initial reasoning over raw data and re-reasoning over previously materialized inferences (``Reasoning Reasoned'' in Table \ref{tab:baselines}), as incremental reasoning is a common workflow. HermiT was excluded due to consistent memory failures. 

The results are presented in Table~\ref{tab:baselines}. RDFlib with owlrl failed to terminate within 180 minutes. Protégé with Pellet demonstrates the fastest reasoning (5.5s raw, 22s reasoned) but requires approximately 5 GB for reasoning and 25 GB for RDF/XML export, limiting scalability. owlready2 reasons less efficient (203.9s raw) and exhausts memory when reasoning over materialized data. GraphDB exhibits consistent performance (1820s raw, 1606s reasoned) with stable memory usage across ontology scales, making it the most reliable alternative to \ac{krrood} for large-scale reasoning.

\ac{krrood} reasoning performance (8.3s raw, 127.9s reasoned) remains competitive while providing native integration with application logic, a capability absent in all other systems. This trade-off is appropriate for applications requiring runtime introspection, procedural reasoning, or tight coupling between knowledge and domain behavior, as required in robotics tasks (see Section~\ref{sec:robot_experiments}).

\subsubsection{Querying}

In a second experiment, we measured the execution times of all queries in the OWL 2 RL profile from OWL2Bench. To ensure consistency, we verified that every framework returned the same set of entities. Protégé failed to terminate on Query 20. To mitigate the impact of the long runtime of Query 20 on the overall assessment, we report the geometric mean across all queries. Q12 was omitted because it would require modeling \textit{ResearchGroup} as a subclass of both Organization and Person, which violates the Liskov Substitution Principle. A \textit{ResearchGroup} cannot safely substitute for a Person.

The results are shown in Table \ref{tab:queries}. SQLAlchemy achieves the fastest performance in most individual cases and maintains the best geometric mean, demonstrating its utility within the KRROOD architecture, especially for large queries. KRROOD with EQL, while not always the fastest, provides fully capable domain objects, which is a requirement for native integration into cognitive systems architectures and is not achieved by any other framework. Conversely, RDFlib, owlready2, and Protégé appear less competitive in this context, leaving GraphDB as the most viable alternative to KRROOD when query speed is the primary concern and having domain objects is not necessary.

\begin{table}[t]
  \centering
   \resizebox{\columnwidth}{!}{%
\begin{tabular}{lll}
\hline
    \textbf{Framework} & 
    \makecell[l]{\textbf{Loading + Reasoning}\\\textbf{Raw}} & 
    \makecell[l]{\textbf{Loading + Reasoning}\\\textbf{Reasoned}} \\
    \hline
KRROOD & $8.32$ & $127.89$\\
RDFLib & $36.011$ & $>10800$ \\
owlready2 & $203.914$ & o. o. m.\\
Protege & $\mathbf{5.465}$ & $\mathbf{22.032}$\\
GraphDB & $1820$ & $1606$\\
\hline
\end{tabular}
  }
  \caption{Loading and reasoning times for different frameworks in seconds. The raw data contains 54,897 statements. The reasoned data contains 1,502,966 statements.
  }
  \label{tab:baselines}
\end{table}

\begin{table*}[t]
  \centering
  \small
\resizebox{\textwidth}{!}{%
\begin{tabular}{llllllll}
\hline
\textbf{Query} & \textbf{Results} & \textbf{SQLAlchemy} & \textbf{GraphDB} & \textbf{EQL} & \textbf{RDFLib} & \textbf{owlready2} & \textbf{Protege}\\
\hline
Q2 & 7421 & $\mathbf{3.78 \pm 0.24}$ & $20.02 \pm 4.74$ & $58.77 \pm 0.96$ & $99.90 \pm 18.08$ & $142.37 \pm 18.19$ & $1620.4 \pm 74.08$\\
Q3 & 55 & $\mathbf{0.58 \pm 0.36}$ & $2.69 \pm 0.58$ & $2.11 \pm 0.20$ & $2.91 \pm 0.84$ & $43.32 \pm 0.75$  & $15.7 \pm 1.418$\\
Q4 & 2486 & $86.22 \pm 250.70$ & $\mathbf{8.67 \pm 1.22}$ & $29.31 \pm 1.40$ & $34.61 \pm 0.73$ & $90.57 \pm 0.48$ & $17.9 \pm 5.238$ \\
Q5 & 20 & $1.33 \pm 0.43$ & $2.72 \pm 0.77$ & $1.52 \pm 0.12$ & $2.30 \pm 0.58$ & $7.15 \pm 0.22$ &  $\mathbf{0.5 \pm 1.581 }$\\
Q7 & 1684 & $\mathbf{1.28 \pm 0.10}$ & $6.65 \pm 0.35$ & $10.01 \pm 0.43$ & $24.65 \pm 0.62$ & $113.44 \pm 1.29$ & $4193.2 \pm 341.577$ \\
Q8 & 6 & $\mathbf{0.54 \pm 0.22}$ & $2.46 \pm 0.33$ & $1.77 \pm 0.06$ & $1.99 \pm 0.12$ & $43.25 \pm 0.77$ & $12.8 \pm 1.229$\\
%Q9 & 0 & $1.21 \pm 0.30$ & $2.19 \pm 0.09$ & $4.34 \pm 0.09$ & $2.03 \pm 0.05$ & $\mathbf{0.49 \pm 0.01}$ & n.a. \\
Q10 & 666 & $\mathbf{0.80 \pm 0.09}$ & $4.63 \pm 1.69$ & $23.75 \pm 0.88$ & $13.01 \pm 0.22$ & $75.16 \pm 0.95$ & $17.1 \pm  1.370$ \\
Q11 & 2422 & $\mathbf{1.52 \pm 0.07}$ & $8.81 \pm 1.55$ & $35.18 \pm 1.49$ & $32.24 \pm 0.66$ & $76.62 \pm 0.38$  & $20.8 \pm 1.317$\\
% Q12 & 2494 & $86.60 \pm 218.74$ & $7.08 \pm 0.28$ & $\mathbf{4.52 \pm 0.39}$ & $29.39 \pm 1.03$ & $85.17 \pm 0.92$ & $18.2 \pm 3.645$\\
%Q13 & 0 & $3.95 \pm 1.47$ & $2.34 \pm 0.29$ & $1.89 \pm 0.61$ & $2.42 \pm 0.73$ & $\mathbf{0.61 \pm 0.17}$ & n.a\\
Q15 & 21 & $\mathbf{0.51 \pm 0.18}$ & $2.41 \pm 0.31$ & $14.30 \pm 0.52$ & $2.37 \pm 0.33$ & $31.13 \pm 1.94$ & $12.5 \pm 0.85$\\
Q16 & 21 & $\mathbf{0.49 \pm 0.11}$ & $2.36 \pm 0.14$ & $1.73 \pm 0.05$ & $2.21 \pm 0.10$ & $27.57 \pm 0.84$  & $12.6 \pm 1.174$ \\
Q19 & 858 & $4.25 \pm 0.19$ & $\mathbf{3.94 \pm 0.21}$ & $4.74 \pm 0.25$ & $11.11 \pm 0.42$ & $49.11 \pm 0.42$  & $5.6 \pm 0.843$\\
Q20 & 1311932 & $\mathbf{626.28 \pm 259.77}$ & $2779.74 \pm 95.42$ & $7926.19 \pm 151.63$ & $28957.92 \pm 456.97$ & $1764.62 \pm 24.80$ & $>60000.0$\\
Q21 & 145 & $18.21 \pm 25.90$ & $30.05 \pm 80.55$ & $61.12 \pm 17.09$ & $63.08 \pm 69.27$ & $\mathbf{11.11 \pm 11.00}$ & $11038.1 \pm 450.084$ \\
Q22 & 106 & $10.45 \pm 1.46$ & $\mathbf{3.15 \pm 0.51}$ & $288.94 \pm 20.58$ & $1625.85 \pm 37.89$ & $6.23 \pm 0.19$ & $554.9 \pm 24.324$ \\
\hline \\
\textbf{Geom. Mean} & --- & $\mathbf{3.39}$ & $8.07$ & $19.72$ & $26.45$ & $49.82$ & $78.33$ \\
\hline
\end{tabular}
}
  \caption{OWL2Bench (RL profile) query performance across frameworks. The results column displays the number of data points returned and is consistent across all frameworks. The framework-specific columns show the time in milliseconds averaged over 10 runs and the standard deviation.}
  \label{tab:queries}
\end{table*}

\subsection{Application: Interactive Robot Task Understanding}
\label{sec:robot_experiments}
\begin{figure}[h]
    \centering
    \includegraphics[width=\linewidth]{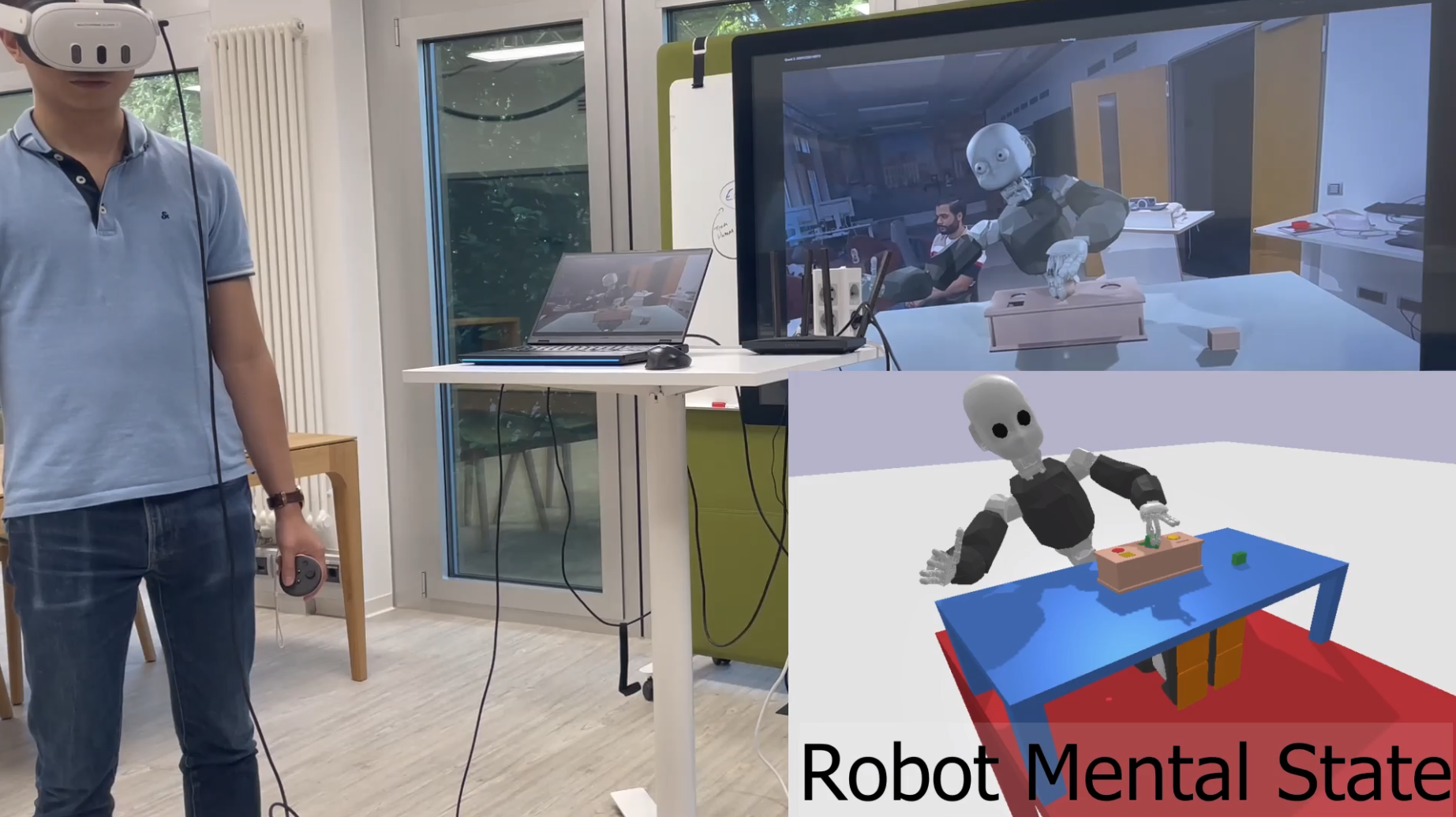}
    \caption{A virtual robot performing an object insertion learned from human demonstration, with the human interacting through a \ac{vr} interface.}
    \label{fig:MultiverseDemo}
\end{figure}

To assess the suitability of \ac{krrood} for \ac{krr} applications in physical intelligence systems, we conducted a series of human-robot interactive task learning experiments. A human demonstrates pick-and-insert actions using a Montessori box, and the robot learns to reproduce the task by observing object–hole insertion patterns and selecting its own actions accordingly (see Fig. \ref{fig:MultiverseDemo}).

Action annotation and reasoning are handled by Segmind, an interactive robot-learning framework based on \ac{krrood}, which represents agents, actions, and interaction relations symbolically to enable logical inference and explainable decision-making. The experiments were implemented in Multiverse\footnote{\url{https://github.com/Multiverse-Framework/Multiverse}} \cite{Multiverse}, which records complete interaction scenarios via a \ac{vr} headset with hand tracking. Visual rendering is performed in Unreal Engine 5, while physics simulation is handled by MuJoCo \cite{todorov2012mujoco}, with all interaction data recorded for analysis.

The interaction follows a demonstration, feedback, and generalization loop. The human first demonstrates a correct object insertion, after which the robot attempts the task based on the observed action. Upon failure, the robot requests a corrective demonstration through audio and infers the object–hole matching constraint. Using this inferred rule, the robot successfully completes the insertion task for the remaining objects.\footnote{A video of the experiment, experiment code, and an interactive simulation environment will be provided online after double-blind review.}

During task execution, Segmind continuously identifies objects and events relevant to the task. The robot infers new task-relevant knowledge using previously learned inference rules stored as \ac{rdr} trees. The \ac{rdr} trees are contextual by design. For example, a single tree may encode the sequence of dynamic events required to recognize a \textit{PickUp} action. Each tree can contain multiple rules corresponding to different situations encountered during the robot’s operation. These rules are incrementally constructed through interaction with the human, who answers the robot’s queries during learning. Reasoning in this framework goes beyond simple rule matching by allowing individual rules to invoke different reasoning mechanisms, such as geometric reasoning (e.g., shapes or holes), spatial reasoning (e.g., containment), and physics-based reasoning. This heterogeneity is supported by the flexible design of the RDR implementation.

Listing~\ref{lst:segmind_rule_tree} illustrates a portion of the RDR tree in Segmind that annotates \textit{PickUp}-related events. A rule detects a \textit{LossOfContactEvent} involving a support object, where contact events are inferred via physics-based collision reasoning in the Multiverse simulation. This leads to a \textit{LossOfSupport} inference. Using a General RDR, the tree is re-evaluated until no new inferences are produced, allowing refinement rules to fire in subsequent passes. As a result, the event is classified as a \textit{PickUpEvent}, unless a refinement rule determines that the object is falling.

\begin{lstlisting}[
  style=clean,
  caption={Part of the RDR tree for event annotation},
  label={lst:segmind_rule_tree},
  emph={Add},
  emphstyle=\color{blue}\bfseries,
  emph={[2]HasType, inference, refinement, variable_from},
  emphstyle={[2]\color{purple}},
  emph={[3]PickUpEvent, FallingEvent, LossOfSupportEvent, LossOfContactEvent, Support},
  emphstyle={[3]\color{teal}\bfseries},
  emph={[4]with},
  emphstyle={[4]\color{brown}}
]
event = variable_from(domain=new_events)
with HasType(event, LossOfContactEvent) 
     & HasType(event.other_body, Support):
  Add(inference(LossOfSupportEvent)(body=event.body, support=event.other_body))
  with HasType(event, LossOfSupportEvent):
    Add(inference(PickUpEvent)(body=event.body))
    with refinement(Equal(event.body.acceleration, GRAVITY_ACC):
      Add(inference(FallingEvent)(body=event.body))
\end{lstlisting}

At any point during execution, queries can be issued about the robot’s current knowledge of the world, and about how specific events were inferred. Such explanations are enabled by two core capabilities of \ac{krrood}: (i) the ability to trace which rules contributed to a given conclusion, and (ii) the ability of \ac{eql} to access domain knowledge represented as class data structures, including the inference rules themselves.

\section{Discussion and Conclusion}
\label{sec:discussion_conclusion}

This paper introduces \ac{krrood}, a Python-native \ac{krr} system built on \ac{ood} principles. The experiments indicate that treating knowledge as a first-class abstraction in Python can bridge the representational gap between \ac{krr} tooling and application-centric software engineering.

\textbf{Capabilities.}
\ac{krrood} enables hybrid AI systems by representing knowledge as Python domain objects that are directly usable by application code. This allows \ac{eql} predicates and \ac{rdr} rule conditions to invoke domain procedures (e.g., geometry, kinematics, simulation) and avoids reliance on external reasoners, while supporting end-to-end introspection and explanation during execution \cite{beetz2025robot,mania2024open,alt2023knowledge,stelter2022open}. Second, the combination of \ac{eql} with ORMatic enables knowledge engineering with \ac{oop} and efficient relational persistence with manageable overheads. \ac{krrood} incurs higher loading costs due to object materialization, but provides competitive reasoning and query capabilities.
% To add (maybe): Competitive in reasoning/querying/loading time to existing KRR implementations that are using OWL.(while also using less resources), Closed world assumption so it will terminate (and can return the expected answers), CC is done natively. RDR's providing a way to incrementally and interactively acquire knowledge.

\textbf{Limitations.}
\ac{eql} makes several principled operational choices (e.g. negation-as-failure and active-domain universal quantification) that align with closed-world assumptions common in physical \ac{ai} applications, but differ from open-world \ac{dl} semantics. Allowing arbitrary Python code in predicates and rules increases practical expressivity but weakens static guarantees such as worst-case runtime and termination. Likewise, Ontomatic preserves most OWL-DL semantics but cannot be fully lossless in general due to modeling mismatches (e.g., non-disjoint sibling classes). Large-scale deployments may incur high memory and latency costs from materializing rich object graphs, though reasoning with \ac{krrood} is orders of magnitude more memory-efficient than OWL-based alternatives.

\textbf{Future work.}
Promising directions include defining stricter \ac{eql} fragments with explicit complexity bounds, as well as  developing hybrid materialization strategies that instantiate only task-relevant subgraphs on demand. \Ac{krrood} will be evaluated on large-scale, long horizon experiments on intelligent robot task planning and control, human-robot task learning and explainable machine learning. 
%\todo{What is the actual future work?}
%Using Krrood for machine learning and explanations? Michael oftens mention, that in order to be able to really explain the behaviour of an agent, you will need the detailed internal state model as a knowledge representation, which with krrood can be enabled 
% LLM-generated knowledge for owl vs krrood
\appendix 

\bibliographystyle{named}
\bibliography{bibliography}

\appendix

\end{document}